# Random Sentences from a Generalized Phrase-Structure Grammar Interpreter

Rick Dale

Department of Psychology, University of Memphis
Memphis, TN, 38152
radale@memphis.edu, http://cia.psyc.memphis.edu/rad/

In numerous domains in cognitive science it is often useful to have a source for randomly generated corpora. These corpora may serve as a foundation for artificial stimuli in a learning experiment (e.g., Ellefson & Christiansen, 2000), or as input into computational models (e.g., Christiansen & Dale, 2001). The following compact and general C program interprets a phrase-structure grammar specified in a text file. It follows parameters set at a Unix or Unix-based command-line and generates a corpus of random sentences from that grammar.

## Command-line arguments

Assuming the sentgen.c source (see below) has been compiled using *cc -lm* into the binary "sentgen," the following is the parameters specified for the interpreter:

> sentgen **filename -e** # -s # -h filename -t filename -v

Required switches are in bold face. The first and required input into the program is a file that contains a phrase-structure grammar description (see below). "-e #" specifies the number of random sentences to be generated, with the randomization seeded by "-s #". "-h filename" will save the selected sentences (as composed of terminals; see below). "-t filename" stores a TLearn-compatible input corpus for a simple-recurrent network. "-v" asks the program to spit random sentences out to stdout.

## The grammar file

The sentence generator ("sentgen") requires an ASCII file that contains a phrase-structure grammar script. The following outlines what is required. Each grammar file must be headed by a sentence node ("highest-ranked" constituent), for example:

```
! comment
S>NP,VP
NP>dog,cat
VP>eats,sleeps
```

Terminal nodes are defined as elements for which no rewrite rule exists. In the above example, dog, cat, eats, sleeps would be chosen randomly (uniformly) for print-out. Probabilities for rewrite are generated by using a period after the constituent, as follows:

```
S>NP,PP.3,VP
NP>dog,cat
PP>P,NP
```

P>on,with
    VP>sleeps,drinks

In the above example, a prepositional phrase (PP) gets printed with .3 probability. You may also include options within nodes as follows:

    S>NP,PP.3,VP_int.3|VP_trans.7
    ...

The final constituent of a sentence here will be an intransitive verb phrase (VP_int) with .3 probability, and transitive verb phrase (VP_trans) with a probability of .7. The program does not require these probabilities to sum to 1. If they do not sum to 1, then if SumP is the sum of the optional probabilities, then with a probability of (1-SumP) the program will print nothing. You may have any number of options:

    S>declarative_s.3|interrogative_s.5|imperative_s.2

Note that the option only applies to elements delimited within commas. In the previous example, transitive and intransitive verbs are the options, and NP and PP have their own respective rewrite probabilities.

## Concluding remarks

This very general specification permits a rich range of phrase-structure grammars for the purposes of experiments or simulations. Also, by iteratively generating corpora, one may use the "-s #" switch to randomize each. Appendices 1 and 2 contain the source code for the program, along with sample grammars.

## Papers using the program

# Appendix 1: Source code

You may also obtain this source code here:

## http://cia.psyc.memphis.edu/downloads/sentgen/.

```
/***********
Generalized Phrase-Structured Grammar Interpreter and Sentence Generator
-------------------------------------------------------------------------
The following code reads in a .grm (grammar) file, parses the phrase-
structure grammar, and generates example sentences.

Written by: Rick Dale, Cornell University
Date last modified: October 18th, 2000 / September 1st, 2001
***********/

#include <stdio.h>
#include <stdlib.h>
#include <string.h>
#include <math.h>

#define TRUE 1
#define FALSE 0

char usage[512] = "\
        \n    sentgen v0 usage: sentgen <grammar file>\
                \n                    [-e(xamples) #]\
                \n                    [-v(erbose)]\
                \n                    [-s(eed) #]\
                \n                    [-o(utput rule array)]\
                \n                    [-h(uman format save to) filename]\
                \n                    [-c(omputer format save to) filename]\
                \n                    [-t(learn file name for .teach/.data) filename]\n\n";

int verbose, seed, i, j, examples, output, total_patterns;
int debugger;
/* grammar file, human file, computer file, tlearn file, tlearn temp file */
char filename[50], h_filename[50], c_filename[50], t_filename_data[50], t_filename_teach[50], t_filename_temp[50];
char first_segment[50];
FILE *hfh, *cfh, *tfh_data, *tfh_teach;
/*
The following contains the label of the node,
e.g. "S", and a list of subnodes (as two-dimensional
array), e.g. "N","V","PP".
*/
struct node {
        char label[255];
        char subnodes[50][255];
} nodes[255];
```

```c
main(int argc, char *argv[]) {

        int current_example;

        total_patterns = 0; /* Counts # of input patterns */

        validate_command_line(argc, argv);

        if (h_filename) hfh = fopen(h_filename, "w");
        if (c_filename) cfh = fopen(c_filename, "w");
        if (t_filename_data) tfh_data = fopen(t_filename_data, "w");
        if (t_filename_teach) tfh_teach = fopen(t_filename_teach, "w");

        load_rule_array(filename);

        if (output) output_rule_array();

        printf("\n");

        /* Each iteration of build_sentences (and its subsequent reiterations)
        builds one example sentence of the grammar */
        for (current_example = 1; current_example <= examples; current_example++) {
                build_sentences(0);
                if (verbose) printf("\n");
                if (hfh) fprintf(hfh, "\n");
                if (cfh) fprintf(cfh, "\n");
        }

        printf("\n");

        if (tfh_teach) fprintf(tfh_teach, "%d %s", total_patterns - 1, first_segment);

        if (hfh) fclose(hfh);
        if (cfh) fclose(cfh);
        if (tfh_teach) fclose(tfh_teach);
        if (tfh_data) fclose(tfh_data);

        if (strlen(t_filename_data)) final_tfc_parse();

        return TRUE;
}
```

```
/**********
This function adds the "localist" line and appends
the number of patterns to the file.  Also, adds pattern
number to the left margin.
**********/
final_tfc_parse(int node_index) {

        FILE *head;
        char line[255];

        /* Print localist head and total patterns */
        head = fopen("vhead", "w");
        fprintf(head, "localist\n");
        fprintf(head, "%d\n", total_patterns);
        fflush(head);
        close(head);

        /* Combine the header with the .data & .teach files */
        sprintf(line, "cat vhead %s > tyzzy; mv tyzzy %s", t_filename_data, t_filename_data);

        system(line);

        sprintf(line, "cat vhead %s > tyzzy; mv tyzzy %s", t_filename_teach, t_filename_teach);

        system(line);
        system("rm -f vhead");

}

/**********
Builds the sample sentences using the global struct array.  By calling
itself recursively, this function cascades down the phrase-structure rules
as defined in the struct array
**********/
build_sentences(int node_index) {
        int step1;
        /*
        If the node is a terminal node, and call this function again. This
        function then cascades down the rewrite rules until it encounters tokens
        */
        for (step1 = 0; strlen(nodes[node_index].subnodes[step1]); step1++) {
                if (is_terminal(nodes[node_index].subnodes[step1])) {
                        build_sentences(is_terminal(nodes[node_index].subnodes[step1]));
                }
                else {
                        /* If the label contains "." then we know options must be computed. */
                        if (strchr(nodes[node_index].subnodes[step1], '.')) {
                                handle_options(nodes[node_index].subnodes[step1]);
                                }
                        else {
                                display_token(node_index);
                                step1 = 999;
                        }
                }
        }
}
```

```
/***********
Handles options/probabilities in any subnode constituent
***********/
handle_options(char option[255]) {
        char labels[10][255];
        float probs[10], rnd_pick, total_rnd;
        int step1, step2;
        char *sub_string, *temp;
        char holder[255];

        /* Make sure labels is initialized for every option */
        for (step1 = 0; step1 <= 9; step1++) strcpy(labels[step1], "\0");

        step1 = 0;

        strcpy(holder, option);

        /* We must first parse all the available options passed to the function */
        sub_string = strtok(holder, "|");

        strcpy(labels[0], sub_string);

        while ((sub_string = strtok(NULL, "|")) != NULL) {
                step1++;
                strcpy(labels[step1], sub_string);
        }

        /* Loop through and get probabilities for the options */
        for (step2 = 0; step2 <= step1; step2++) {
                sub_string = strtok(labels[step2], ".");
                temp = sub_string;
                sub_string = strtok(NULL, ".");
                probs[step2] = (float) atoi(sub_string) / power(10, strlen(sub_string));
        }

        /* Generate a random number between 0 and 1, loop until that number falls
        inside added probabilities */
        rnd_pick = drand48();
        total_rnd = 0;

        for (step1 = 0; strlen(labels[step1]); step1++) {
                total_rnd += probs[step1];
                if (rnd_pick <= total_rnd) {
                        build_sentences(is_terminal(labels[step1]));
                        return TRUE;
                }
        }
}

/***********
Power function: 10^x
***********/
power(int base, int exponent) {
        int step1;
        double result;
        result = 1;
        for (step1 = 1; step1 <= exponent; step1++) result = result * base;
        return result;
}
```

```c
/***********
Displays a terminal node (token) randomly using appropriate list of tokens in rule set
***********/
display_token(int node_index) {
        int no_tokens;
        double token_selected;
        char *token, *input_units;
        char holder[255];
        no_tokens = ubound(node_index);

        /* It is assumed that the tokens listed are uniformly random */
        token_selected = drand48() / (1 / (double) no_tokens);
        /* Display the token by the token index chosen by the random number */
        strcpy(holder, nodes[node_index].subnodes[(unsigned int) token_selected]);
        token = strtok(holder, "}");
        if (verbose) printf("%s ", token);

        if (hfh) fprintf(hfh, "%s ", token);

        /* If we've chosen to build the TLearn file, then we need to parse the
        values delimited by "+" and print out the numbers in the file name specified.
        The resulting file will be parsed later. */
        if (tfh_data) {
                token = strtok(NULL, "}");
                input_units = strtok(token, "+");

                fprintf(tfh_data, "%s", input_units);

                /* If it's the first, don't print in the teach (it has to predict the next), and if it's
                the first, store the first segment to print out at the end */
                if (total_patterns > 0)  {
                        fprintf(tfh_teach, "%d %s", total_patterns - 1, input_units);
                }
                else {
                        sprintf(first_segment, "%s", input_units);
                }

                while ((input_units = strtok(NULL, "+")) != NULL) {
                        fprintf(tfh_data, ",%s", input_units);
                        /* Similarly here -- if it's the first, load first_segment for later printing into teach file */
                        if (total_patterns > 0)  {
                                fprintf(tfh_teach, ",%s", input_units);
                        }
                        else {
                                strcpy(holder, first_segment);
                                sprintf(first_segment, "%s,%s", holder, input_units);
                        }
                }
                fprintf(tfh_data, "\n");
                if (total_patterns > 0)  fprintf(tfh_teach, "\n");
        }
        total_patterns++;

        if (cfh) fprintf(cfh, "%s ", nodes[node_index].label);
}
```

```c
/***********
Determines # of subnodes in the rule array for a specific node
***********/
ubound(int node_index) {
        i = 0;
        while (strlen(nodes[node_index].subnodes[i])) i++;
        return i;
}

/***********
Determines whether certain label is terminal node, or has constituents,
and returns the node_index if so
***********/
is_terminal(char label[255]) {
        for (j = 0; strlen(nodes[j].label); j++) {
                if (strcmp(nodes[j].label, label) == 0) return j;
        }
        return FALSE;
}

/***********
Makes sure the command-line arguments are valid, and initialize variables
***********/
validate_command_line(int argc, char *argv[]) {
        if (argc == 1) {
                fprintf(stderr, "%s", usage);
                exit(0);
        }
        strcpy(filename, argv[1]);
        for (i = 2; i < argc; i++) {
        if (!strncmp(argv[i], "-v", 2)) verbose = TRUE;
        if (!strncmp(argv[i], "-o", 2)) output = TRUE;
        if (!strncmp(argv[i], "-s", 2)) seed = atoi(argv[++i]);
        if (!strncmp(argv[i], "-e", 2)) examples = atoi(argv[++i]);
        if (!strncmp(argv[i], "-h", 2)) strcpy(h_filename, argv[++i]);
        if (!strncmp(argv[i], "-c", 2)) strcpy(c_filename, argv[++i]);
        if (!strncmp(argv[i], "-t", 2)) {
                sprintf(t_filename_data, "%s.data", argv[++i]);
                        sprintf(t_filename_teach, "%s.teach", argv[i]);
        }

        }

    srand48(seed);
        return TRUE;
}
```

```c
/***********
Loads the phrase-structure rule information from file specified in the
command-line, and ignores any line beginning with "!" -- the comment
marker
***********/
load_rule_array(char filename[50]) {
        FILE *fh;
        char line[255];

        fh = fopen(filename, "r");
        if (!fh) {
                fprintf(stderr, "\n   Error opening file.  Please check that the file exists,\n   and is in the current directory.\n
%s", usage);
                exit(0);
        }
        j = 0;
        while (fgets(line, 255, fh)) {
                if (line[0] != '!' && line[0] != '\n') {
                        parse_line(line, j);
                        j++;
                }
        }
        fclose(fh);
        return TRUE;
}

parse_line(char line[255], int j) {
        char *sub_string;
        line = strtok(line, "\n"); /* Trim the last line character */
        sub_string = strtok(line, ">"); /* Get the labe and save it */
        strcpy(nodes[j].label, sub_string);
        sub_string = strtok(NULL, ">");
        sub_string = strtok(sub_string, ",");
        strcpy(nodes[j].subnodes[0], sub_string); /* Loop through delimited subnodes, and save them */
        for (i = 1; (sub_string = strtok(NULL, ",")) != NULL; i++) {
                strcpy(nodes[j].subnodes[i], sub_string);
        }
}

/***********
Displays total rule set.  Used for debugging
***********/
output_rule_array() {
        printf("\n");
        for (i = 0; strlen(nodes[i].label); i++) {
                printf(" %d) %s--> ", i + 1, nodes[i].label);
                for (j = 0; strlen(nodes[i].subnodes[j]); j++) {
                        if (strlen(nodes[i].subnodes[j + 1])) {
                                printf("%s,", nodes[i].subnodes[j]);
                        }
                        else {
                                printf("%s\n", nodes[i].subnodes[j]);
                        }
                }
        }
}
```

Appendix 2: Grammar examples

1)

! You can create comments in the .grm file using !
S>NP,VP
NP>det,N
det>the,a
N>cat,dog
VP>jumps,eats

2)

S>NP,PP.33,V_intransitive.3 | V_transitive.7
NP>cat,dog
PP>P,NP
P>on,with
V_intransitive>sleeps,swims
V_transitive>V_trans,NP
V_trans>kicks,eats